%% file: main.tex
\documentclass[letterpaper, 10 pt, conference]{ieeeconf}  % Comment this line out
\IEEEoverridecommandlockouts                              % This command is only
\let\labelindent\relax

\usepackage[utf8]{inputenc} % allow utf-8 input
\usepackage[T1]{fontenc}    % use 8-bit T1 fonts
\usepackage{hyperref}       % hyperlinks
\usepackage{url}            % simple URL typesetting
\usepackage{booktabs}       % professional-quality tables
\usepackage{amsfonts}       % blackboard math symbols
\usepackage{nicefrac}       % compact symbols for 1/2, etc.
\usepackage{microtype}      % microtypography
\usepackage{xcolor}         % colors
\usepackage{graphicx}
\usepackage{subcaption}
\usepackage{amsmath,amssymb,amsthm,bm} % commented for \proof already defined error

\newtheorem*{lemma*}{Lemma}
\newtheorem*{claim*}{Claim}

\usepackage{mathtools, nccmath}
\usepackage{multicol,multirow,amsmath,bm}
\usepackage[ruled,vlined,noline,noend]{algorithm2e}
\usepackage{blindtext}
\usepackage{soul}

\theoremstyle{plain}
\newtheorem{theorem}{Theorem}

\newtheorem{lemma}[theorem]{Lemma}
\newtheorem{corollary}[theorem]{Corollary}
\theoremstyle{definition}
\newtheorem{definition}{Definition}
\newtheorem{assumption}{Assumption}

\usepackage{tikz}
\usepackage{anyfontsize}
\usetikzlibrary{arrows,matrix,positioning,decorations.pathreplacing,calc}
\usepackage{hhline}
\allowdisplaybreaks

\newcommand{\policy}{$\mathsf{MOCHA}$~}
\newcommand{\policyns}{$\mathsf{MOCHA}$}

\usepackage{comment}
\usepackage{enumitem}
\usepackage{wrapfig}
\usepackage{ulem}

\setenumerate{label=\arabic*)}% global settings

\input{symbols_commands}

\title{\LARGE \bf
Enabling Pareto-Stationarity Exploration in Multi-Objective Reinforcement Learning: A Multi-Objective Weighted-Chebyshev Actor-Critic Approach
}

\author{Fnu Hairi$^{1}$,Jiao Yang$^{2}$, Tianchen Zhou$^{2}$, Haibo Yang$^{3}$, Chaosheng Dong$^{2}$, \\Fan Yang$^{2}$, Michinari Momma$^{2}$, Yan Gao$^{2}$, Jia Liu$^{4}$%
\thanks{$^{1}$Department of Computer Science, University of Wisconsin-Whitewater, Whitewater, WI, USA.
         Email: {\tt\small hairif@uww.edu}}%
\thanks{$^{2}$Amazon, Seattle, USA.
        Email: {\tt\small\{jaoyan, tiancz,chaosd,fnam,michi,yanngao\}@amazon.com}}
\thanks{$^{3}$Department of Computing and Information Sciences, Rochester Institute of Technology, Rochester, NY, USA.
        Email: {\tt\small hbycis@rit.edu}}%
\thanks{$^{4}$Department of Electrical and Computer Engineering, The Ohio State University, Columbus, OH, USA.
        Email: {\tt\small liu@ece.osu.edu}}%
}

\begin{document}

\maketitle
\thispagestyle{empty}
\pagestyle{empty}

%%%%%%%%%%%%%%%%%%%%%%%%%%%%%%%%%%%%%%%%%%%%%%%%%%%%%%%%%%%%%%%%%%%%%%%%%%%%%%%%
\begin{abstract}
In many multi-objective reinforcement learning (MORL) applications, being able to systematically explore the Pareto-stationary solutions under multiple non-convex reward objectives with theoretical finite-time sample complexity guarantee is an important and yet under-explored problem.
This motivates us to take the first step and fill the important gap in MORL. Specifically, in this paper, we propose a \uline{M}ulti-\uline{O}bjective weighted-\uline{CH}ebyshev \uline{A}ctor-critic (\policyns) algorithm for MORL, which judiciously integrates the weighted-Chebychev (WC) and actor-critic framework to enable Pareto-stationarity exploration systematically with finite-time sample complexity guarantee.
Sample complexity result of \policy algorithm reveals an interesting dependency on $p_{\min}$ in finding an $\epsilon$-Pareto-stationary solution, where $p_{\min}$ denotes the minimum entry of a given weight vector $\bm{p}$ in WC-scalarization. By carefully choosing learning rates, the sample complexity for each exploration can be $\tilde{\mathcal{O}}(\epsilon^{-2})$. 
Furthermore, simulation studies on a large KuaiRand offline dataset, show that the performance of \policy algorithm significantly outperforms other baseline MORL approaches.
\end{abstract}

%%%%%%%%%%%%%%%%%%%%%%%%%%%%%%%%%%%%%%%%%%%%%%%%%%%%%%%%%%%%%%%%%%%%%%%%%%%%%%%%
\input{intro}
\input{related_work}
\input{model}
\input{actor_critic}

\input{analysis}
\input{experiment}

\input{conclusion}
%%%%%%%%%%%%%%%%%%%%%%%%%%%%%%%%%%%%%%%%%%%%%%%%%%%%%%%%%%%%%%%%%%%%%%%%%%%%%%%%

\bibliographystyle{IEEEtran} % or your chosen style
\bibliography{refs,refs1}

\end{document}

%% file: symbols_commands.tex
\renewcommand{\d}{\mathbf{d}}

\newcommand{\F}{\mathbf{F}}
\newcommand{\g}{\mathbf{g}}

\newcommand{\J}{\mathbf{J}}
\newcommand{\K}{\mathbf{K}}

\newcommand{\p}{\mathbf{p}}

\renewcommand{\r}{\mathbf{r}}

\newcommand{\w}{\mathbf{w}}

\newcommand{\x}{\mathbf{x}}

\newcommand{\1}{\mathbf{1}}

%% file: intro.tex
\section{Introduction} \label{sec: intro}
Multi-objective systems \cite{HayRduBar_22} have gained significant attention due to their applicabilities in many real-world applications. For example, on commercial platforms like Booking.com, in addition to the overall ratings for satisfaction, hotels receive various customer ratings for subcategories such as \textit{value-for-money}, \textit{comfort}, and \textit{cleanliness}. These ratings potentially provide a more nuanced recommendation strategy than the traditional overall rating-based recommendation system. From the perspective of a decision-maker (whether a client or a recommender system), the goal is to develop decision-making strategies that maximize all these ratings to deliver an ideal service experience. Despite these multiple ratings seemingly providing more insights about the hotels, they sometimes can conflict. For instance, hotels with high cleanliness ratings often cost more and, leading to low value-for-money ratings. Consequently, Pareto optimality is more suitable solution concepts in this case, where overall balanced solutions can be provided for a further decision-making. Another example is short-term video recommender system on video platforms. In general, the recommender system aim to engage users more on the platform by maximizing the pleasant experience while minimizing the negative ones. Specifically, on Kuaishou \cite{CaiXueZha_23} platform, it considers multiple objectives such as "WatchTime", "Likes", "Forward", "Comments", "Dislikes" to optimize.

Reinforcement learning (RL) \cite{SutBar_18} provides a learning framework where agents learn policies through experience by trial and error. Although the definition for RL is over scalar rewards, it can be naturally extended to vectorized reward settings \cite{ZhoHaiYan_24}, known as multi-objective reinforcement learning (MORL). The key differences between single objective RL and MORL lie in their objective goals and solution concepts. In single-objective RL, the goal is to learn optimal policies that maximizes the long-term accumulated rewards as follows
\begin{align}
\max_{\pi} \mathbb{E}[\gamma r_1 + \gamma^{2} r_2 + \cdots + \gamma^{t} r_t + \cdots] \nonumber
\end{align}
where $\pi$ denotes a candidate policy in the policy space, $\gamma$ the discount factor and the expectation is subject to usual caveats about appropriate distribution.
Similarly, in MORL, the goal extends to finding a policy that maximizes the following vectorized objective
\begin{align}
&\max_{\pi} \mathbb{E}[\gamma \mathbf{r}_1 + \gamma^{2} \mathbf{r}_2 + \cdots + \gamma^{t} \mathbf{r}_t + \cdots] \nonumber %\\
\end{align}
where $\mathbf{r}_t$ denotes the vectorized reward. 

A policy $\pi$ is Pareto optimal if it is not dominated by any other policy $\pi'$. However, in general, there are more than one Pareto optimal solutions for multi-objective problems. Pareto front (PF) is a set that includes all Pareto optimal solutions. One of the main research endeavors is to characterize PF systematically. Generally speaking, when the objective functions are non-convex or the PF is non-connected, it is shown challenging to characterize PF \cite{DanDvuGas_22,YanLiuLiu_23,KimdeW_06,JadShaSin_24}. In this paper, we consider the concept of Pareto stationarity front (PSF), which consists of all Pareto stationary solutions (see the definitions later). By definition, PSF is a superset of PF. The goal of the paper is to develop an algorithm that characterizes the set of PSF in MORL problems.

In this paper, we propose a \uline{M}ulti-\uline{O}bjective weighted-\uline{CH}ebyshev \uline{A}ctor-critic (\policyns) method by drawing inspirations and insights from the MORL and multi-objective optimization (MOO) literature.
More specifically, to enable systematic Pareto-stationarity front exploration with low sample complexity in MORL, our proposed \policy method takes advantage of approach of multiple temporal-difference (TD) learning in the critic component and multi-gradient-descent algorithmic (MGDA) techniques in the actor component, originally proposed in \cite{ZhoHaiYan_24}, then judiciously integrates the weighted-Chebyshev (WC).
The rationale behind our approach is three-fold:
(i) Combining the strengths of value-based and policy-based RL approaches, the actor-critic framework has been shown to offer state-of-the-art performance in RL;
(ii) in the MOO literature, it has been shown that an optimal solution under the WC-based scalarization approach (also known as hypervolume scalarization) provably achieves the Pareto front even when the Pareto front is non-convex~\cite{ZhaGol_20};
and (iii) for MOO problems, the MGDA method is an efficient approach for finding a Pareto-stationary solution~\cite{Des_12}.
Finally, the above connections leads us to generalize the WC and actor-critic framework to our \policy method for MORL.

{\bf Main Challenges:}
However, to show that \policy enjoys systematic Pareto-stationarity exploration with provable low sample complexity remains highly non-trivial due to the fact:
In the MOO literature, WC- and MGDA-based techniques are developed with very different goals in mind, facilitating Pareto-front exploration and achieving Pareto-stationarity, respectively. 
To date, it remains unclear how to combine them to achieve systematic Pareto-stationarity exploration with finite-time convergence and low sample complexity simultaneously even for general MOO problems, not to mention generalizing them to the more specially structured MORL problems and the associated theoretical performance analysis.
Indeed, to our knowledge, there is no such result in the literature on integrating WC- and MGDA- techniques for designing MORL policies.

%% file: related_work.tex
\section{Related Work} \label{sec: related_work}
In this section, we provide an overview of related work in multi-objective reinforcement learning. 

Without learning framework, multi-objective optimization \cite{Mie_99,Des_12} has been studied extensively with various problem settings, solution concepts and corresponding approaches. Notably, the weighted-Chebyshev formulation and multiple-gradient descent algorithm used in this paper can be traced back to their standard adoption in MOO \cite{Mie_99} and \cite{Des_12}, respectively. 
MGDA can be viewed as an extension of the standard gradient descent method to MOO, which dynamically performs a linear combination of all objectives' gradients in each iteration to identify a common descent direction for all objectives.
\cite{FliVazVic_19} and \cite{liu2021stochastic} established finite-time convergence of $\mathcal{O}(1/T)$ to Pareto stationarity point for MGDA and Stochastic MGDA respectively. 
Also, the finite-time convergence rate of MGDA has recently been established under different MOO settings, including convex and non-convex objective functions \cite{liu2021stochastic,FerSheLiu_22} and decentralized data \cite{YanLiuLiu_23}, etc. \cite{XiaBanJi_23} proposed a weight/direction vector oriented stochastic gradient descent algorithm in MOO.

MORL (also referred to as multi-criteria reinforcement learning) dates back to at least \cite{GabKalSze98}, where a Q-learning based algorithm is proposed for constraint setting. In solving Pareto stationary solution, \cite{CheDuXia_21} proposed an actor-critic where in the critic, it minimizes the target loss and in the actor, it uses policy gradient to update deterministic stationary policy. Subsequently, \cite{CaiXueZha_23} proposed a two-staged constrained actor-critic algorithm, in which, among the multiple objectives, one is selected as a primary objective and the remaining ones are considered as constraints.  \cite{ZhoHaiYan_24} proposed an MGDA based actor-critic algorithm that finds a Pareto stationary solution. Finite-time and sample complexity results have been provided with an $M$-independence property. We note that in this work, we adopt a similar framework but judiciously incorporate WC formulation \cite{LinLiuZha_24,MomDonLiu_22,qiu2024traversingparetooptimalpolicies}, which enables the systematic exploration. 

In terms of exploring PF, \cite{LinZheLi_19} proposed a linear scalarization based multi-objective learning to approximate Pareto front. However, to the best of our knowledge, our work is the first attempt in exploring the Pareto stationarity front.
%Multi-task learning \cite{Car_97,GhoBen_96,SenKol_18} generally refers to the framework where similarly the learning for related tasks can be bootstrapped via sharing data among different tasks than otherwise independently. Here, in the context of reinforcement learning, the similarity shares within the transition kernel of the environment. However, the reward entries within each reward vector can be arbitrary; in other words, they can be similar or competing.

%% file: model.tex
\section{MORL Problem Formulation}
In this section, we introduce the problem formulation and preliminaries of MORL problems.

{\bf 1) Multi-Objective Markov Decision Process (MOMDP) \cite{ZhoHaiYan_24,roijers2018multi}:}
MOMDP is a stochastic process characterized by the following tuple $(\mathcal{S},\mathcal{A},P,\mathbf{r},\mathbf{\gamma})$. $\mathcal{S}$ and $\mathcal{A}$ denote the state space, action space respectively. $P: \mathcal{S}\times\mathcal{A} \xrightarrow{} \mathcal{S}$ denotes the transition kernel. $\mathbf{r}: \mathcal{S}\times\mathcal{A} \xrightarrow{} [0,r_{\max}]^{M}$, is an $M$-dimensional vector rewards, where $r_{\max}>0$ is a reward upper bound constant. $\mathbf{\gamma}$ denotes the discount vector, where $\gamma^{i}\in (0,1)$ denotes the discount factor for objective $i\in [M]$ and $[M]:=\{1,\cdots,M\}$.

The key differences between MOMDP and single-objective MDP \cite{QiuYanYe_21,XuWanLia_20,SutBar_18} are vectorized reward for $M$-objective and potentially different discount factors for objectives. For simplicity, we consider finite state and action spaces.

We consider a universal stationary policy $\bm{\pi}(\cdot|s)$ for all $s\in \mathcal{S}$, in this paper. In other words, the agent maintains a single universal policy to balance all $M$-objectives. Furthermore, we consider the policy $\bm{\pi}$ to be parameterized by a $d$-dimensional parameter $\bm{\theta}$, i.e. $\bm{\pi}_{\bm{\theta}}$. Moreover, we assume that $\bm{\pi}_{\bm{\theta}}$ is continuously differential with respect to $\bm{\theta}$, which is a necessary condition for applying policy gradient approach. A typical parameterized policy can be soft-max functions. Next, we impose the assumption on the underlying Markov chains.
\begin{assumption}
For all $\bm{\theta}\in \mathbb{R}^{d}$, the state Markov chain $\lbrace s_t\rbrace_{t\geq 0}$ induced by the policy $\pi_{\bm{\theta}}$ is irreducible and aperiodic.
\label{ass:mdp}
\end{assumption}
The above assumption implies that there's a unique stationary distribution for the state Markov chain with transition matrix $P_{\bm{\theta}}(s'|s)=\sum_{a\in\mathcal{A}}\pi_{\bm{\theta}}(a|s)\cdot P(s'|s,a), \forall s,s'\in\mathcal{S}$. This is a standard assumption adopted in many literature \cite{QiuYanYe_21,XuWanLia_20,roijers2018multi,HaiLiuLu_22,ZhaYanLiu_18}.

{\bf 2) Learning Objective and Pareto Solution Concepts:}
For each objective $i \in [M]$, the objective function is the accumulated discounted reward in infinite horizon, as in conventional RL.
$$J^{i}(\bm{\theta}):=\mathbb{E} [ \sum_{t=1}^{\infty} (\gamma^{i})^t r^{i}_{t}(s_t, a_t) ],$$
where expectation is taken over state-action visitation occupancy measure given an initial distribution and $\gamma^{i}\!\in\!(0,1)$ is the discount factor associated with objective $i$.
The goal of MORL is to find an optimal policy $\pi_{\bm{\theta^*}}$ with parameters $\bm{\theta}^*$ to jointly maximize all objective's long-term rewards in the sense of Pareto-optimality (to be defined next).
Specifically, 
we want to learn a policy $\pi_{\bm{\theta}}$ that maximizes the following vector-valued objective:
\begin{equation}
\max_{\bm{\theta}\in\mathbb{R}^{d}} \J(\bm{\theta}) := [J^{1}(\bm{\theta}), \ldots, J^{M}(\bm{\theta})]^{\top}.
\label{eqn:MORL}
\end{equation}

As mentioned in Section~\ref{sec: intro}, due to the fact that the objectives in MORL are conflicting in general, the more appropriate and relevant learning goal and optimality notions in MORL are the Pareto-optimality and the Pareto front, which are defined as follows:
\begin{definition}[(Weak) Pareto-Optimal Policy and (Weak) Pareto Front]
    We say that a policy $\pi_{\bm{\theta}}$ dominates another policy $\pi_{\bm{\theta}'}$ if and only if $J^i(\bm{\theta})\geq J^i(\bm{\theta}'), \forall i\in[M]$ and $J^i(\bm{\theta})> J^i(\bm{\theta}'), \exists i\in[M]$. 
    A policy $\pi_{\bm{\theta}}$ is Pareto-optimal if it is not dominated by any other policy.
    A policy $\pi_{\bm{\theta}}$ is weak Pareto-optimal if and only if there does not exist a policy $\pi_{\bm{\theta'}}$ such that $J^i(\bm{\theta'}) > J^i(\bm{\theta}), \forall i\in[M]$.
    Moreover, the image of all (weak) Pareto-optimal policies constitute the (weak) Pareto front.
\end{definition}
In plain language, a Pareto-optimal policy identifies an equilibrium where no reward objective can be further increased without reducing another reward objective, while a weak Pareto-optimal policy characterizes a situation where no policy can simultaneously improve the values of all reward objectives (i.e., ties are allowed).
However, since MORL problems are often non-convex in practice (e.g., using neural networks for policy modeling or evaluation), finding a weak Pareto-optimal policy is NP-hard.
As a result, finding an even weaker Pareto-stationary policy is often pursued in practice. 
Formally, let $\nabla_{\bm{\theta}} J^{i}(\bm{\theta})$ represent the policy gradient (to be defined later) direction of the $i$-th objective with respect to $\bm{\theta}$.
A Pareto-stationary policy is defined as follows:
\begin{definition}[Pareto-Stationary Policy] \label{Def:PS}
A policy $\pi_{\bm{\theta}}$ is said to be Pareto-stationary if there exists no common ascent direction $\d \in \mathbb{R}^{d}$ such that $\d^{\top} \nabla_{\bm{\theta}} J^{i}(\bm{\theta}) >0$ for all $i\in [M]$, where 
    \begin{align}
    \nabla_{\bm{\theta}}\J(\bm{\theta})=
    \left[\begin{matrix}
    \nabla_{\bm{\theta}} J^{1}(\bm{\theta}) &
    \nabla_{\bm{\theta}} J^{2}(\bm{\theta}) &
    \cdots &
    \nabla_{\bm{\theta}} J^{M}(\bm{\theta})
    \end{matrix}\right] \in \mathbb{R}^{d\times M}.  \nonumber
\end{align} 
\end{definition}
Since MORL is a special-structured MOO problem, it follows from the MOO literature that Pareto stationarity is a necessary condition for a policy to be Pareto-optimal\cite{Des_12}.
Note that in convex MORL settings where all objective functions are convex functions, Pareto-stationary solutions imply Pareto-optimal solutions.

\begin{definition}[(Pareto-Stationarity Front]
The image of all Pareto-Stationary policies constitute the Pareto Stationarity front.
\end{definition}
In this paper, we propose a weighted-Chebyshev formulation, inspired by Lemma 1 in Section \ref{sec:policy} \cite{qiu2024traversingparetooptimalpolicies}, which takes advantage of MGDA approach to systematically explore the Pareto stationarity front(PSF). In order to represent the PSF better, WC formulation requires the exploration using a well-represented exploration set $\mathcal{P} = \{\bm{p}_1,\cdots,\bm{p}_n\}$ in parallel. In this paper, we recommend the exploration set $\mathcal{P}$ should uniformly cover unit angular weight vectors similar to \cite{LinZheLi_19}.

%% file: actor_critic.tex
\section{\policyns: Algorithm Design and Theoretical Results} \label{sec:policy}

In this section, we propose \policy method for solving MORL problems.
As mentioned in Section~\ref{sec: intro}, our \policy algorithm is motivated by two key observations:
(i) actor-critic approaches combine the strengths of both value-based and policy-based approaches to offer the state-of-the-art RL performances;
and (ii) an optimal solution under the WC-based scalarization provably achieves the Pareto front even for non-convex MOO problems.
In what follows, we will first introduce some preliminaries of \policy in Section~\ref{sec:moac_prelim}, which are needed to present our \policy algorithmic design in Section~\ref{subsec:pf-moac-alg}. 
Lastly, we will present the finite-time Pareto-stationary convergence and sample complexity results of \policy in Section~\ref{subsec:pf-moac-theories}.

\subsection{Preliminaries for the Proposed \policy Algorithm} \label{sec:moac_prelim}
Similar to single-objective actor-critic methods, the critic component in \policy evaluates the current policy by applying TD learning for all objectives. 
However, the novelty of \policy stems from the actor component, which applies policy-gradient updates by judiciously combining 1) WC-scalarization and 2) MGDA-style updates motivated from the MOO literature.

{\bf 1) Weighted-Cheybshev Scalarization:}
The WC-scalarization is a scalarization technique in MOO that converts a vector-valued objective into a scalar-valued optimization problem, which is more amenable for algorithm design.
Specifically, let $\Delta_M$ represent the $M$-dimensional probability simplex.
For a multi-objective loss minimization problem $\min_{\x} \F(\x) := [f_1(\x),\ldots,f_{M}(\x)]^{\top} \in \mathbb{R}_{+}^{M}$, the WC-scalarization with a weight vector $\p \in \Delta_M$ is defined in the following min-max form:
\begin{align} \label{eqn:WC_def}
\mathsf{WC}_{\p}(\F(\cdot)) := \min_{\x} \max_i \{ p_i f_i(\x)\}_{i=1}^{M} = \min_{\x} \| \p \odot \F(\x) \ \|_{\infty},
\end{align}
where $\odot$ denotes the Hadamard product. 
The use of WC-scalarization in our \policy algorithmic design is inspired by the following fact in MOO~\cite{Golovin2020RandomHS,qiu2024traversingparetooptimalpolicies}:
\begin{lemma}(Proposition 4.7 in \cite{qiu2024traversingparetooptimalpolicies}) \label{lem_WC}
A solution $\x^*$ is weakly Pareto-optimal to the problem $\min_{\x} \F(\x)$ if and only if $\x^* \in \arg\min_{\x} \mathsf{WC}_{\p}(\F(\x))$ for some $\p \in \Delta_M$.    
\end{lemma}
Lemma~\ref{lem_WC} suggests that, by adopting WC-scalarization in MORL algorithm design (since MORL is a special class of MOO problems), we can systematically obtain all weakly Pareto-optimal policies (i.e., exploring the weak Pareto front) by enumerating the WC-scalarization weight vector $\p$ if the WC-scalarization problem can be solved optimally.
As will be seen later, this motivates our \policy design in Section~\ref{subsec:pf-moac-alg}.

{\bf 2) Policy Gradient for MORL:}
Since the actor component in our \policy algorithm is a policy-gradient approach, it is necessary to formally define policy gradients for MORL. 
Toward this end, we first define the advantage function for each reward objective $i\in [M]$:
$\text{Adv}^{i}_{\bm{\theta}}(s,a)=Q^{i}_{\bm{\theta}}(s,a)-V^{i}_{\bm{\theta}}(s)$, where $Q^{i}_{\bm{\theta}}(s,a)$ and $V^{i}_{\bm{\theta}}(s)$ are the Q-function and value function for the $i$-th objective under policy $\bm{\pi}_{\bm{\theta}}$. 
Let $\bm{\psi}_{\bm{\theta}}(s,a):=\nabla_{\bm{\theta}}\log \pi_{\bm{\theta}}(a|s)$ be the score function for state-action pair $(s,a)$.
Then, policy gradient for the $i$-th objective is computed as follows:
\begin{lemma}[Policy Gradient Theorem \cite{ZhoHaiYan_24}]
Let $\pi_{\bm{\theta}}:\mathcal{S}\times\mathcal{A}\to[0,1]$ be any policy and $J^{i}(\bm{\theta})$ be the accumulated reward function for the $i$-th objective. 
Then, the policy-gradient of $J^{i}(\bm{\theta})$ with respect to policy parameter $\bm{\theta}$ is:
$\nabla_{\bm{\theta}}J^{i}(\bm{\theta})=\mathbb{E}_{s\sim d_{\bm{\theta}}(\cdot),a\sim\pi_{\bm{\theta}}(\cdot|s)} [\bm{\psi}_{\bm{\theta}}(s,a)\cdot \textup{Adv}^{i}_{\bm{\theta}}(s,a)]$, where $d_{\bm{\theta}}(\cdot)$ is the state visitation measure under policy $\bm{\pi}_{\bm{\theta}}$.
\label{the: pol_gra}
\end{lemma}

{\bf 3) Function Approximation:}
To achieve finite-time convergence result for \policy, we adopt linear approximations for value function approximations. The value function for objective $i\in [M]$ is approximated by a linear function. In other words, $V^i(s) \approx \bm{\phi}(s)^{\top}\w^i, i\in[M]$, where $\w^i\in\mathbb{R}^{\tilde{d}}$ with $\tilde{d}\le|\mathcal{S}|$ and $\tilde{d}\in \mathbb{R}$. $\bm{\phi}(s)\in\mathbb{R}^{\tilde{d}}$ is the feature mapping associated with state $s\in\mathcal{S}$ and we use $\Phi\in\mathbb{R}^{|\mathcal{S}|\times \tilde{d}}$ to represent the feature matrix. We impose the following assumption on feature matrix.
\begin{assumption}
\label{assum:LFA}
$\Phi$ is bounded and full rank.
\label{ass:feature}
\end{assumption}
Without loss of generality, we further assume that $\|\phi(s)\|\le 1$ for all $s\in\mathcal{S}$. Assumption~\ref{ass:feature} is standard in the RL literature (e.g., \cite{TsiVan_99,ZhaYanLiu_18,QiuYanYe_21,ZhoHaiYan_24}), is an attempt to deal with RL problems with large state-action space, i.e. $\tilde{d}\ll|\mathcal{S}|$.

\subsection{The Proposed \policy Algorithm Framework} \label{subsec:pf-moac-alg}

With the preliminaries in Section~\ref{sec:moac_prelim}, we are in a position to present our \policy algorithm.
For ease of exposition, we will structure our \policy algorithm design in two main derivation steps.

\textbf{Step 1) Multiple-TD Learning in the Critic Component:}
We note that the multiple-TD learning was first proposed in \cite{ZhoHaiYan_24}. We briefly describe the component for the completeness of the algorithmic presentation.
As stated in Assumption~\ref{ass:feature}, the critic component (i.e., policy evaluation) in \policy maintains value-function approximation parameters $\w^i$ for each objective $i\in[M]$.
For the current policy $\pi_{\bm{\theta}_t}$, the critic component in \policy updates the value function parameters $\w^i_k, i\in[M]$ in parallel via TD learning with mini-batch Markovian samples.
The TD-error $\delta_{k,\tau}^{i}$ for objective $i$ in iteration $k$ using sample $\tau$ can be computed as:
\begin{equation}
    \hspace{.5in} \delta^{i}_{k,\tau} = r^{i}_{k,\tau}+\gamma^{i}\bm{\phi}^{\top}(s_{k,\tau+1})\w^{i}_k - \bm{\phi}^{\top}(s_{k,\tau})\w^{i}_k.
\label{eq:critic_disc}
\end{equation}

Subsequently, each parameter $\w^{i}$ is updated in a batch fashion in parallel using the following TD-learning step: $\w^{i}_{k}= \w^{i}_{k-1}+(\beta/D) \sum_{\tau=1}^{D}\delta^{i}_{k,\tau}\cdot\bm{\phi}(s_{k,\tau})$.
Once the critic component executes $N$ rounds, the parameters $\{\w^{i}\}_{i\in[M]}$ can be used in the actor component for policy evaluation.

\textbf{Step 2) The WC-MGDA-Type Policy Gradient in the Actor Component:} 

As mentioned earlier, the actor component in \policy is a ``multi-gradient'' extension of the policy gradient approach in MORL, which determines a {\em common policy improvement direction} for all reward objectives by dynamically weighting the individual policy gradients.
Toward this end, we will further organize the common policy improvement direction derivations in two key steps as follows:

{\em Step 2-a) WC-Guided Common Policy Improvement Direction:}
First, we compute a dynamic weighting vector $\hat{\bm{\lambda}}^*_t$ in each iteration $t$ that balances two key aspects: 1) find a common policy improvement direction based on multi-TD learning to converge to a Pareto-stationary solution; and 2) follow the guidance of a WC-scalarization weight vector $\p$.
To adopt an MGDA-type policy improvement update in \policy, we first convert the original MORL reward maximization problem in Eq.~\eqref{eqn:MORL} to the following logically equivalent  ``regret minimization'' problem with respect to the Pareto front:
\begin{align} \label{eq:MORL_equiv}
     \min_{\bm{\theta}\in\mathbb{R}^{d}}& \left(\J_{\mathrm{ub}}^{*}-\J(\bm{\theta})\right) \nonumber \\
     :&=\Big[J_{\mathrm{ub}}^{1,*}-J^1(\bm{\theta}), J_{\mathrm{ub}}^{2,*}-J^2(\bm{\theta}), \ldots, J_{\mathrm{ub}}^{M,*}-J^M(\bm{\theta}) \Big]^{\top}, 
\end{align}
where $J_{\mathrm{ub}}^{i,*}$ is an estimated upper bound of $J^{i,*}:=\max_{\bm{\theta}\in\mathbb{R}^{d}} J^{i}(\bm{\theta})$ (i.e., the optimal value of the $i$-th objective under single-objective RL).
The rationale behind using $\mathbf{J}_{\mathrm{ub}}^{*}$ in \eqref{eq:MORL_equiv} is to ensure that the polarity of the reformulated problem is conformal to the standard use of WC-scalarization in MOO.
Note that, regardless of the choice of the $\mathbf{J}_{\mathrm{ub}}^{*}$-estimation, there is always a 1-to-1 mapping between the Pareto fronts between Problems~\eqref{eqn:MORL} and \eqref{eq:MORL_equiv}.
Hence, using the WC-scalarization to explore the Pareto front of Problem~\eqref{eq:MORL_equiv} is logically equivalent to exploring the Pareto front of Problem~\eqref{eqn:MORL}, and the tightness of the $\mathbf{J}_{\mathrm{ub}}^{*}$-estimation is not important.

Next, since Problem~\eqref{eq:MORL_equiv} is in the standard MOO form, according to \cite{Des_12}, the MGDA approach for Problem~\eqref{eq:MORL_equiv} can be written as:
\begin{equation} \label{eqn:reform_MGDA}
\min_{\bm{\lambda}} \|\K \bm{\lambda} \|^{2} \quad \text{s.t.} \quad \mathbf{1}^{\top}\bm{\lambda}=1, \,\, \bm{\lambda}\in\mathbb{R}_{+}^{M},
\end{equation}
where $\K:= \sqrt{\mathbf{G}^{\top}\mathbf{G}}$ and and $\mathbf{G}$ is the gradient matrix of $\J_{\mathrm{ub}}^{*}-\J(\bm{\theta})$.
On the other hand, following Eq.~\eqref{eqn:WC_def}, the WC-scalarization of Eq.~\eqref{eq:MORL_equiv} with a given weight vector $\p$ is: $\min_{\bm{\theta}\in\mathbb{R}^{d}} \| \p \odot \left(\J_{\mathrm{ub}}^{*}-\J(\bm{\theta})\right) \|_{\infty}$,
which can be reformulated as follows by introducing an auxiliary variable $\rho$:
\begin{align} \label{eqn:WC_reform}
\min_{\rho\in \mathbb{R},\bm{\theta}\in\mathbb{R}^{d}} \rho \quad \text{s.t.} \quad \p \odot \left(\J_{\mathrm{ub}}^{*}-\J(\bm{\theta})\right)\le \rho \mathbf{1}. 
\end{align}
By the KKT stationarity condition on $\rho$ and $\bm{\theta}$ and associating Lagrangian dual variables $\bm{\lambda} \in \mathbb{R}_{+}^{M}$, it can be readily verified that the Wolfe dual problem of Eq.~\eqref{eqn:WC_reform} can be written as~\cite{MomDonLiu_22}:
\begin{align} \label{eqn:Wolfe_Dual}
\max_{\bm{\lambda},\bm{\theta}} \bm{\lambda}^{\top}(\p\odot \left(\J_{\mathrm{ub}}^{*}-\J(\bm{\theta})\right)), \nonumber \\
\text{ s.t. } \mathbf{K}_{\p}\bm{\lambda}=0, \,\, \mathbf{1}^{\top}\bm{\lambda}=1, \,\, \bm{\lambda}\in\mathbb{R}_{+}^{M}, \,\, \bm{\theta}\in\mathbb{R}^{d},
\end{align}
where $\bf{K}_p:=\text{diag}(\sqrt{\p})\sqrt{G^{\top}G}\text{diag}(\sqrt{\p})$.
Since the condition $\bf{K}_p\bm{\lambda}=0$ may not be satisfied at all iterations in an algorithm, we incorporate the minimization of  $\| \K_{\p}\bm{\lambda}\|^2$ in \eqref{eqn:Wolfe_Dual} using a parameter $u>0$ to balance the trade-off with the objective $\bm{\lambda}^{\top}(\p\odot \left(\J_{\mathrm{ub}}^{*}-\J(\bm{\theta})\right))$ to yield:
\begin{align} \label{eqn:WC-MGDA}
\min_{\bm{\lambda},\bm{\theta}} \| \K_{\p} \bm{\lambda} \|^2 - u \bm{\lambda}^{\top}(\p\odot \left(\J_{\mathrm{ub}}^{*}-\J(\bm{\theta})\right)) \quad  \nonumber \\
\text{ s.t. }\quad \mathbf{1}^{\top} \bm{\lambda}=1, \,\, \bm{\lambda}\in\mathbb{R}_{+}^{M}, \bm{\theta}\in\mathbb{R}^{d}.
\end{align}
Now, comparing \eqref{eqn:WC-MGDA} with \eqref{eqn:reform_MGDA} and \eqref{eqn:Wolfe_Dual}, it is clear that solving for $\bm{\lambda}$ in Problem~\eqref{eqn:WC-MGDA} under the current $\bm{\theta}$-value yields a $\bm{\lambda}$-weighting of the gradients of $(\J_{\mathrm{ub}}^{*}-\J(\bm{\theta}))$, which achieves a balance between Pareto-front exploration and Pareto-stationarity induced by WC and MGDA, respectively.
Moreover, upon fixing a $\bm{\theta}$-value, solving for $\bm{\lambda}$ in Problem~\eqref{eqn:WC-MGDA}
is a convex quadratic program (QP), which can be efficiently solved similar to the standard MGDA~\cite{Des_12}.
In iteration $t$, let $\hat{\bm{\lambda}}_t^*$ be the solution obtained from solving Problem~\eqref{eqn:WC-MGDA} under current policy parameter $\bm{\theta}_t$.
To mitigate the cumulative systematic bias resulting from $\bm{\lambda}_t$-weighting, we show that one can update $\bm{\lambda}_t$ by using a momentum-based approach \cite{ZhoZhaJia_22,ZhoHaiYan_24} with momentum coefficient $\eta_t\in[0,1)$ as follows:
\begin{equation} \label{eq:lambda}
    \bm{\lambda}_t = (1-\eta_t)\bm{\lambda}_{t-1}+\eta_t\hat{\bm{\lambda}}^*_t.
\end{equation}
Next, with the obtained $\bm{\lambda}_t$ from \eqref{eq:lambda}, we can update policy parameters $\bm{\theta}$ by conducting a gradient-descent-type update in \eqref{eqn:WC-MGDA} as follows: $\bm{\theta}_{t+1} = \bm{\theta}_t -\alpha \mathbf{G}_t (\p \odot \bm{\lambda}_t)$ with step size $\alpha>0$.

{\em Step 2-b) Policy Gradient Computation for Individual Reward Objective:}
Although we have derived the WC-MGDA-type update in Step~2-a, it remains to evaluate the gradient matrix $\mathbf{G}$ of $(\J_{\mathrm{ub}}^{*}-\J(\bm{\theta}))$.
Note that $\J_{\mathrm{ub}}^{*}$ is a constant, each column $\g_t^i$ in $\mathbf{G}$ is equal to the negative policy gradient  
of each reward objective $i$.
To compute $\g_t^i$, %from Assumption~\ref{assum:LFA}, 
the actor component starts with sampling and TD-error computations.
First, from Lemma~\ref{the: pol_gra}, we compute the score function in the $l$-th actor step as follows:
\begin{align}
\bm{\psi}_{t,l}:=\nabla_{\bm{\theta}}\log \pi_{\bm{\theta}_t}(a_{t,l}|s_{t,l}). \label{eq: score}
\end{align}
Next, similar to the critic component, the actor computes the TD-error for objective $i$ at time $t$ using sample $l$ can be computed as follows:

\begin{equation}
    \hspace{.5in} \delta^{i}_{t,l}= r^{i}_{t,l}+\gamma^{i}\bm{\phi}^{\top}(s_{t,l+1})\w^{i}_t-\bm{\phi}^{\top}(s_{t,l})\w^{i}_t.
\label{eq:actor_disc}
\end{equation}

With the score function in \eqref{eq: score} and the TD-error in \eqref{eq:actor_disc}, one can compute the individual policy gradient as $\g_t^i = -{1\over B}\sum_{l=1}^{B}\delta_{t,l}^i\cdot\bm{\psi}_{t,l}$ following Lemma \ref{the: pol_gra}.

In conclusion, we summarize the full \policy in Algorithm~\ref{alg: pf-moac}.

\begin{algorithm*}[t!]
\caption{The \policy Algorithm.}
    \SetKwInOut{Input}{Input}
    \SetKwInOut{Output}{Output}
    \SetKwFor{ParFor}{for}{do in parallel}{end for}
    \Input{Initial State $s_0$, Initial Policy Parameter $\bm{\theta}_1$, Feature Matrix $\Phi$, Discount Factors $\{\gamma^{i}\}_{i\in[M]}$, Initial Critic Parameters $\{\w^{i}_0\}_{i\in [M]}$, Exploration/Weight Vector $\p$, Momentum Coefficients $\{\eta_t\}_{t\in [T]}$, Actor Step Size $\alpha$, Actor Iteration $T$, Actor Batch Size $B$, Critic Step Size $\beta$, Critic Iteration $N$, Critic Batch Size $D$}
    \BlankLine

    \For{$t=1,\cdots,T$}{
        \begin{minipage}[t]{0.95\linewidth}
        \begin{multicols}{2}
        \ul{\textbf{Critic Component:}} 
        
        \For{$k=1,\cdots,N$}{
        $s_{k,1}=s_{k-1,D}$ (when $k=1$, $s_{1,1}=s_0$)
        
        \For{$\tau=1,\cdots,D$}{
            execute action $a_{k,\tau}\sim\pi_{\bm{\theta}_t}(\cdot|s_{k,\tau})$,\\
            observe state $s_{k,\tau+1}$, reward $\r_{k,\tau+1}$
            
            \ParFor{$i\in [M]$}{
                update $\delta^i_{k,\tau}$ by Eq.~(\ref{eq:critic_disc})
            }
        }
        
        \ParFor{$i\in [M]$}{
          TD update:\\ $\w^{i}_{k}= \w^{i}_{k-1}+\frac{\beta}{D}\sum_{\tau=1}^{D}\delta^{i}_{k,\tau}\cdot\bm{\phi}(s_{k,\tau})$
          }
        }
        
        \ParFor{$i\in [M]$}{
        denote $\w^{i}_{t}= \w^{i}_{k}$
        }
        \columnbreak
        \ul{\textbf{Actor Component:}}
        
        \For{$l=1,\cdots,B$}{
            execute action $a_{t,l}\sim\pi_{\bm{\theta}_t}(\cdot|s_{t,l})$,\\
            observe state $s_{t,l+1}$, reward $\r_{t,l+1}$
            
            \ParFor{$i\in [M]$}{
                update $\bm{\psi}_{t,l}$ by Eq.~\eqref{eq: score},
                \\update $\delta^i_{t,l}$ by Eq.~(\ref{eq:actor_disc})
            }
        }
        
        \ParFor{$i\in[M]$}{
            $\g_t^i = -{1\over B}\sum_{l=1}^{B}\delta_{t,l}^i\cdot\bm{\psi}_{t,l}$
        }
        Solve for $\hat{\bm{\lambda}}^*_t$ in Problem (\ref{eqn:WC-MGDA}) under current $\bm{\theta}_t$;\\
        Update $\bm{\lambda}_t$ by Eq.~(\ref{eq:lambda});\\
        Update $\g_t = \bm{G}_t(\p\odot\bm{\lambda}_t)$;\\
        Update policy: $\bm{\theta}_{t+1}= \bm{\theta}_t-\alpha\cdot \g_t$
        \end{multicols}
        \end{minipage}
    }   
    \BlankLine
    \BlankLine
    \Output{$\bm{\theta}_{\hat{T}}$ with $\hat{T}$ chosen uniformly random from $\{1,\cdots,T\}$}
    \label{alg: pf-moac}
\end{algorithm*}

%% file: analysis.tex
\subsection{Theoretical Performance of \policy}
\label{subsec:pf-moac-theories}

In this section, we analyze \policyns's convergence to a Pareto-stationary solution and the associated sample complexity of the \policy for any given weight vector $\p$. 
%Due to space limitations, we relegate all proofs to the Appendix.
For finite-time Pareto-stationary convergence analysis, instead of using the original definition in Defition~\ref{Def:PS}, it is more convenient to use the following equivalent near-Pareto stationarity characterization
defined as follows~\cite{Des_12,SenKol_18,YanLiuLiu_23,ZhoHaiYan_24}:
\begin{definition}($\epsilon$-Pareto Statioinary Point)
    For a given $\epsilon >0$, a solution $\bm{\theta}$ is $\epsilon$-Pareto stationary if there exists $\bm{\lambda}\in\mathbb{R}_{+}^M$ satisfying $\bm{\lambda}\geq \bm{0}$, $\1^{\top}\bm{\lambda}=1$, such that $\|\nabla_{\bm{\theta}}\J(\bm{\theta})\bm{\lambda}\|_2^2\leq\epsilon$.\footnote{We use $\|\cdot\|_2$ to denote $\ell_2$ norm.}
\end{definition}
Next, we state the following assumptions needed for Pareto-stationary convergence analysis: 

\begin{assumption}
(a) For any parameter $\bm{\theta}\in \mathbb{R}^{d}$ and state-action pair $(s,a)\in\mathcal{S}\times\mathcal{A}$, $\|\bm{\psi}_{\bm{\theta}}(s,a)\|_2\le C$ for some $C>0$;
(b) For any two policy parameters $\bm{\theta},\bm{\theta}'\in \mathbb{R}^{d}$ and $\forall i\in[M]$, $\|\nabla_{\bm{\theta}}J^{i}(\bm{\theta})-\nabla_{\bm{\theta}}J^{i}(\bm{\theta}')\|_2\le L\|\bm{\theta}-\bm{\theta}'\|_2$ for some $L>0$.
\label{ass:Lip_bou}
\end{assumption}
In Assumption~\ref{ass:Lip_bou}, Part (a) imposes the score function to be uniformly bounded for all policy and state-action pair and Part (b) imposes the gradient of each objective function is Lipschitz with respect to the policy parameter via a common constant $L$. 
These assumptions are standard and has been adopted in the analysis of the single-objective actor-critic RL algorithms in \cite{QiuYanYe_21,XuWanLia_20} and MORL in \cite{ZhoHaiYan_24}. 
For discounted reward setting, both items can be guaranteed by choosing common policy parameterizations \cite{XuWanLia_20,GuoHuZha_21}.

We let $\zeta_{\text{approx}}:= \max_{i\in[M]}\max_{\bm{\theta}}\mathbb{E}[|V^i(s) - V_{\w^{i,*}}^i(s)|^2]$ represent the approximation error of the critic component, which is zero if the ground-truth value functions $V^{i}(\cdot)$, $\forall i\in [M]$, are in the linear function class;
otherwise, $\zeta_{\text{approx}}$ is non-zero due to the expressivity limit of the critics.

We now state our main convergence theorem of \policy to a neighborhood of a Pareto-stationary point for any given exploration vector $\p$ as follows:
\begin{theorem}
Under Assumptions \ref{ass:mdp}-\ref{ass:Lip_bou}, set the actor and critic step sizes as $\alpha={1\over 3L}$ and $\beta$ a sufficiently small constant. For any momentum coefficient sequence $\{\eta_t\}_{t=1}^{T}$ and vector $\p$ with minimum entry $p_{\min}>0$, the iterations generated by Algorithm~\ref{alg: pf-moac} satisfy the following finite-time Pareto-stationary convergence error bound:
\begin{align*}
     &\mathbb{E}\big[ \|\nabla_{\bm{\theta}}\J(\bm{\theta}_{\hat{T}})\bm{\lambda}_{\hat{T}}\|^2_2 \big]\leq \mathcal{O}\left(\frac{1}{T}(1+\frac{2\sum_{t=1}^{T}\eta_t}{p^{2}_{\min}})\right)+\mathcal{O}(\cfrac{1}{B})  \nonumber \\
     & +\mathcal{O}(\max_{j\in [M], t\in [T]}\mathbb{E}\left[\left\|\w^j_t - \w^{j,*}_t\right\|_2^2\right])+ \mathcal{O}(\zeta_{\mathrm{approx}}),
\end{align*}
where $\hat{T}$ is sampled uniformly among $\{1,\cdots, T\}$.
\label{thm:moac1}
\end{theorem}

Theorem~\ref{thm:moac1} suggests that the convergence depends on the interplay between momentum coefficient sequence $\{\eta_t\}_{t=1}^{T}$ and the minimum entry $p_{\min}$ of the WC-scalarization weight vector $\p$: 1) The larger $\sum_{t=1}^{T}\eta_t$ or the smaller $p_{\min}$, \policy requires larger iteration $T$ to Pareto-stationary convergence; 2) By letting $\eta_t = \frac{p^{2}_{\min}}{t^{2}}$, the first term on the right-hand-side of Theorem~\ref{thm:moac1} will be $\mathcal{O}(\frac{1}{T})$. As a result of the above insight, the order-wise convergence result matches weight-free Pareto stationary convergence in \cite{ZhoHaiYan_24} and also single-objective RL convergence to stationary policy in \cite{XuWanLia_20}. We remark that the step size for critic can be the same as in single-objective counterpart in \cite{XuWanLia_20}.

\begin{corollary}
Under the same conditions as in Theorem \ref{thm:moac1}, for any $\epsilon>0$,
by setting $\eta_t\!=\!p^{2}_{\min}/t^{2}$, 
$T=\Theta(1/\epsilon), \mathbb{E}[\|\w^i_t - \w^{i,*}_t\|_2^2]= \mathcal{O}(\epsilon), \forall i\in[M], t\in[M]$, and $B=\Theta(1/\epsilon)$, we have
    $\mathbb{E}\big[ \| \nabla_{\bm{\theta}}\J(\bm{\theta}_{\hat{T}})\bm{\lambda}_{\hat{T}}\|^2_2 \big]\leq \mathcal{O}(\epsilon) + \mathcal{O}\big(\zeta_{\mathrm{approx}}\big)$,
with total sample complexity of $\mathcal{O} (\epsilon^{-2}\log{(\epsilon^{-1})} )$.

\label{corollary:conv_rate}
\end{corollary}

Note that Theorem~\ref{thm:moac1} and Corollary~\ref{corollary:conv_rate} show the convergence rate of \policy are {\em independent} of the number of objectives $M$ as in \cite{ZhoHaiYan_24} even in the presence of weight vector $\p$. When $\p$ is all-one vector, the results in Theorem \ref{thm:moac1} and Corollary \ref{corollary:conv_rate} recovers those of \cite{ZhoHaiYan_24}. In other words, \policy is a more general algorithm than MOAC in \cite{ZhoHaiYan_24}.

%% file: experiment.tex
\section{Experiments}
\label{sec: exp}

In this section, we empirically evaluate \policy and compare it with other related state-of-the-art methods on a large-scale real-world dataset.
%Due to space limitations, we present the main experimental results here and relegate the full experimental setting details to the Appendix.

\textbf{1) Dataset:} We leverage a large-scale recommendation logs dataset from short video-sharing mobile app Kuaishou\footnote{\url{https://kuairand.com/}} as in \cite{CaiXueZha_23,ZhoHaiYan_24}.
The dataset includes 
multiple reward signals, such as ``Click'', ``Like'', ``Comment'', ``Dislike'', ``WatchTime'' and etc.
The statistics for the dataset is summarized in Table~\ref{data}.
Here, a state corresponds to the event that a video is watched by a user and is represented by concatenating user and video features; an action corresponds to recommending a video to a user.

\begin{table}[h]
\caption{Statistics for Dataset}
\centering
\resizebox{.5\linewidth}{!}{
\setlength{\tabcolsep}{3pt}
\begin{tabular}{cccccc}
\toprule
\multicolumn{6}{l}{State: $1218$ \quad Action: 150} \\
\toprule
&\multicolumn{5}{c}{Reward} \\
\cmidrule(lr){2-6}
& Click & Like & Comment & Dislike & WatchTime \\
\midrule
Amount & $254940$ & $5190$ &$1438$&$213$&$199122$\\
\midrule
Density &$55.25\%$&$1.125\%$&$0.312\%$&$0.046\%$&$43.15\%$\\
\bottomrule
\end{tabular}
}
\label{data}
\end{table}

\begin{table*}[h]
\caption{Comparison of \policy with baseline methods given a weight vector.}
\centering
\resizebox{.75\textwidth}{!}{
\begin{tabular}{cccccc}
\toprule
Objective & Click$\uparrow$ & Like$\uparrow$(e-2) & Comment$\uparrow$(e-3) & Dislike$\downarrow$(e-4) & WatchTime$\uparrow$ \\
weights & $0.2$ & $0.2$ & $0.2$ & $0$ & $0.4$ \\
\midrule
Behavior-Clone & $0.534$ & $1.231$ & $3.225$ & $2.304$ & $1.285$ \\
\midrule
TSCAC & $0.549$  & $1.328$  & $2.877$  & $1.177$  & $1.365$  \\
      & $2.75\%$ & $7.88\%$ &$-10.80\%$ &$-48.92\%$ &$6.23\%$\\
\midrule
SDMGrad & $0.543$  & $1.279$  & $3.136$  & $1.166^*$  & $1.329$ \\
     & $1.79\%$ &$3.87\%$ & $-2.77\%$ &$-49.41\%^*$ &$3.46\%$ \\
\midrule
MOAC & $0.541$  & $1.312$  & $3.266^*$  & $1.486$  & $1.307$ \\
     & $1.30\%$ &$6.57\%$ & $1.27\%^*$ &$-35.5\%$ &$1.71\%$ \\
\toprule
\policy  & $\bm{0.555}$  & $\bm{1.329}$  & $3.092$  & $1.339$  & $\bm{1.375}$ \\
{\bf (Ours)}  & $\bm{3.97\%}$ & $\bm{7.96\%}$ & $-4.12\%$ & $-41.88\%$ &$\bm{7.00\%}$ \\
\bottomrule
\end{tabular}
}
\label{exp_table}
\end{table*}

\begin{figure*}[t!]
    \centering
    \begin{subfigure}[t]{0.32\textwidth}
        \centering
        \includegraphics[trim=0 50 0 50, width=\textwidth]{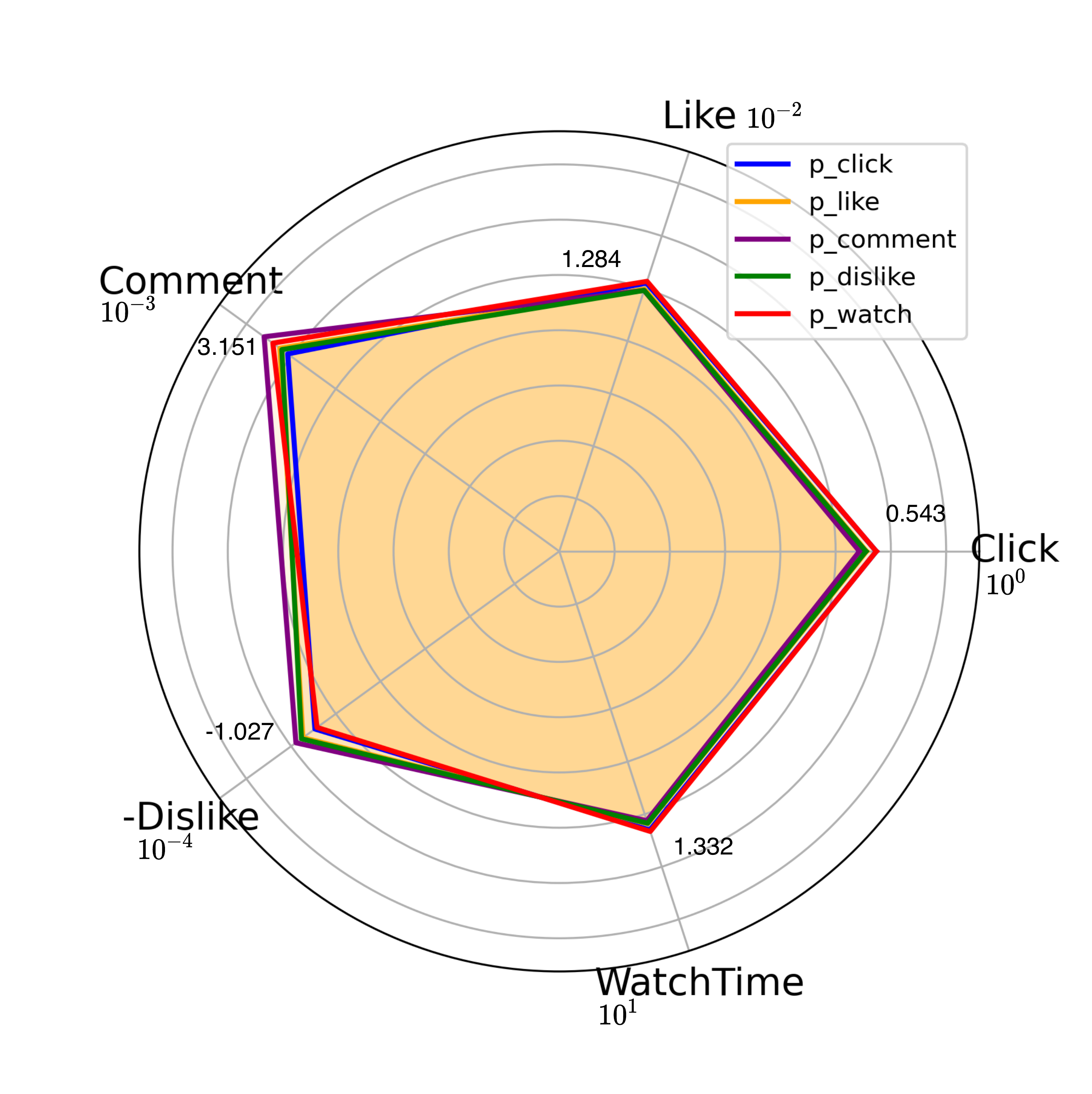}
        \caption{SDMGrad.}
    \label{exp_figa}
    \end{subfigure}%
    ~
    \begin{subfigure}[t]{0.32\textwidth}
        \centering
        \includegraphics[trim=0 50 0 50, width=\textwidth]{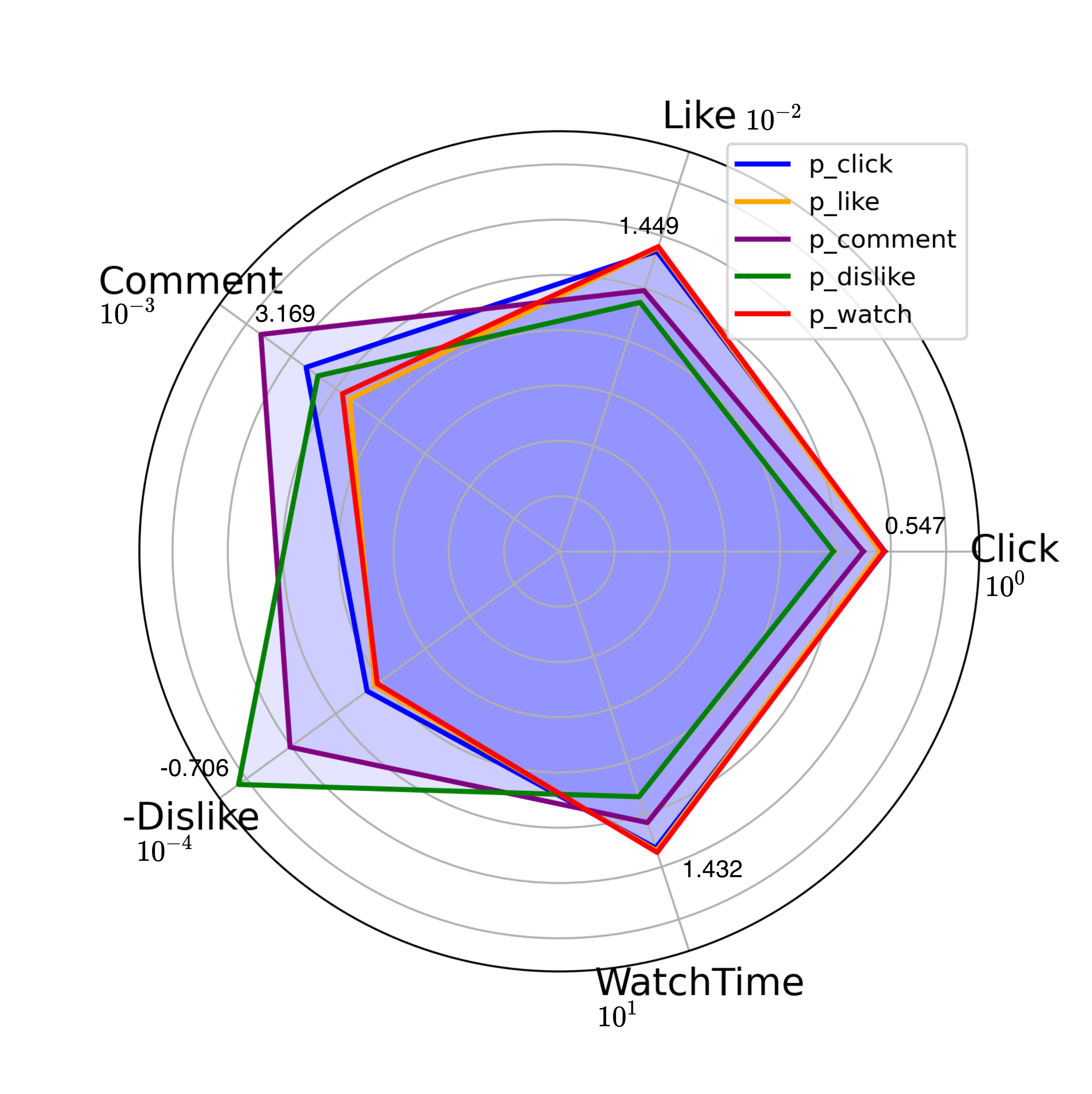}
        \caption{\policy.}
    \label{exp_figb}
    \end{subfigure}%
    ~ 
    \begin{subfigure}[t]{0.32\textwidth}
        \centering
        \includegraphics[trim=0 50 0 50, width=\textwidth]{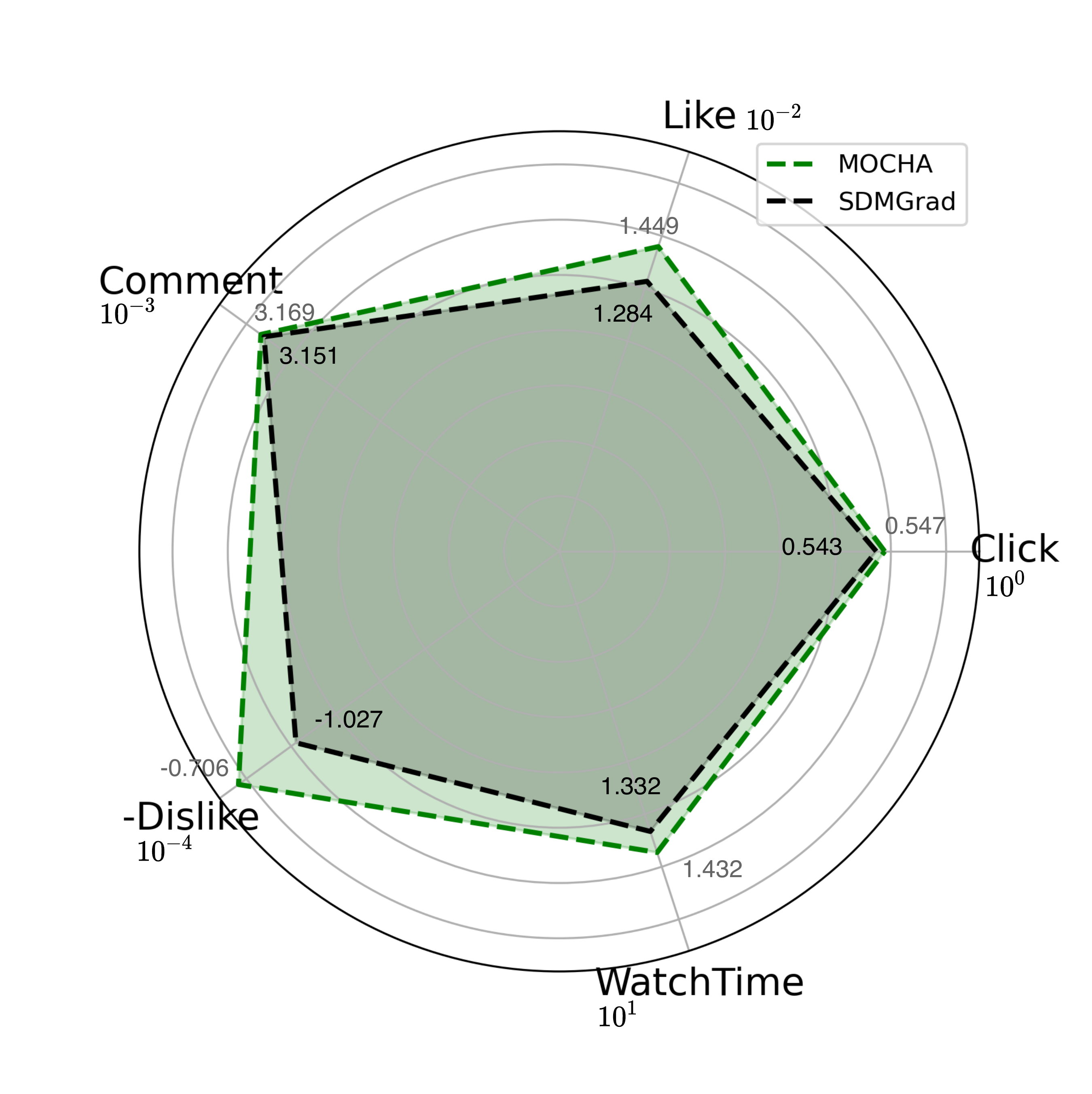}
        \caption{Footprints of exploration.}
    \label{exp_figc}
    \end{subfigure}
    \caption{Comparison of \policy and SDMGrad with five one-hot weight vectors.}
\vspace{-10pt}
\label{exp_fig}
\end{figure*}

\textbf{2) Baselines:}
In this experiment, we leverage the following state-of-the-art methods as baselines:
\begin{list}{\labelitemi}{\leftmargin=1.5em \itemindent=-0.0em \itemsep=-.2em}
\item \textbf{Behavior-Clone}: A behavior-cloning policy $\pi_{\beta}$ that is trained through supervised learning to learn the recommendation policy in the dataset.
\item \textbf{TSCAC} \cite{CaiXueZha_23}: An $\xi$-constrained actor-critic approach that 
optimizes a single objective (i.e., ``WatchTime''), while treating other objectives as constraints bounded by some $\xi>0$.
\item \textbf{SDMGrad} \cite{XiaBanJi_24}: A weight/direction vector $\p$ oriented stochastic gradient descent algorithm, which is shown to find an $\epsilon$-accurate Pareto stationary point. We note that this algorithm has the most potential to explore various Pareto stationary solutions, due to flexibility of adjusting weight vector $\p$.
\item \textbf{MOAC} \cite{ZhoHaiYan_24}: An actor-critic algorithm that aims to find a Pareto Stationarity policy. We note that MOAC doesn't explore PSF, but rather finds an arbitrary Pareto Stationary solution.
\end{list}
Due to the dataset being a static offline dataset,
we adapt \policy and baseline algorithms to off-policy setting. 
We adopt normalized capped importance sampling (NCIS), a standard evaluation approach for off-policy RL algorithms \cite{zou2019reinforcement,ZhoHaiYan_24} to evaluate all methods.
The definition of NCIS for each $i\in[M]$ for a given policy $\bm{\pi}$ is as follows:
\begin{equation*}
    \text{NCIS}^{i}(\bm{\pi}) = \cfrac{\sum_{s,a\in D}\text{CIS}(s,a)r^{i}(s,a)}{\sum_{s,a\in D}\text{CIS}(s,a)}
\end{equation*}
where $\text{CIS}(s,a) = \min\bigg\lbrace C, \cfrac{\bm{\pi}(a\mid s)}{\bm{\pi}_{\beta}(a\mid s)} \bigg\rbrace$, $D$ is the dataset, $C$ is a positive constant to cap the important sampling, and $\bm{\pi}_{\beta}$ is the behavior policy. By definition, a larger NCIS score implies a better performance for a corresponding objective. All methods are initialized with same critic and actor parameters. In addition, initial policies for all methods are set to be the same policy that performs worse than the behavior policy $\bm{\pi}_{\beta}$.

\textbf{3) Results and Observations:} 
We summarize the performance of all methods based on a given weight vector in Table~\ref{exp_table}.
We set the weight vector $\p$ to be $(0.2, 0.2, 0.2, 0, 0.4)^{\top}$ for ``Click'', ``Like'', ``Comment'', ``Dislike'', and ``WatchTime'', respectively.
Note that TSCAC does not require a weight vector since it only optimizes “WatchTime”. %as much as it can.
From Table~\ref{exp_table}, we observe that \policy outperforms SDMGrad, TSCAC and MOAC in three out of five objectives, which are ``Click'', ``Like'', and ``WatchTime''. MOAC and SDMGrad perform best in ``Comment'' and ``Dislike'' objectives, respectively, whereas \policy performs the third in both objectives among the five approaches. The above observation implies that \policy is performing the best overall. 

In Fig.~\ref{exp_fig}, we set the weight vector to be one-hot vectors with ``Click'', ``Like'', ``Comment'', ``Dislike'', and ``WatchTime'' as the only objective, respectively. Fig.~\ref{exp_fig} only illustrate the comparison between \policy and SDMGrad (since TSCAC cannot explore Pareto front).
All figures are plotted in the same scale.
Comparing Fig.~\ref{exp_figa} and Fig.~\ref{exp_figb}, we observe that i) \policy is optimizing the corresponding objectives more than those in SDMGrad; ii) among all the weight vector directions, \policy possesses a larger footprint in the radar chart than SDMGrad (see Fig.~\ref{exp_figc}), which shows that \policy 
has a better Pareto stationarity exploration performance.

\begin{figure}[t!]
    \centering
    \begin{subfigure}[t]{0.25\textwidth}
        \centering
        \includegraphics[trim=0 50 0 50, width=\textwidth]{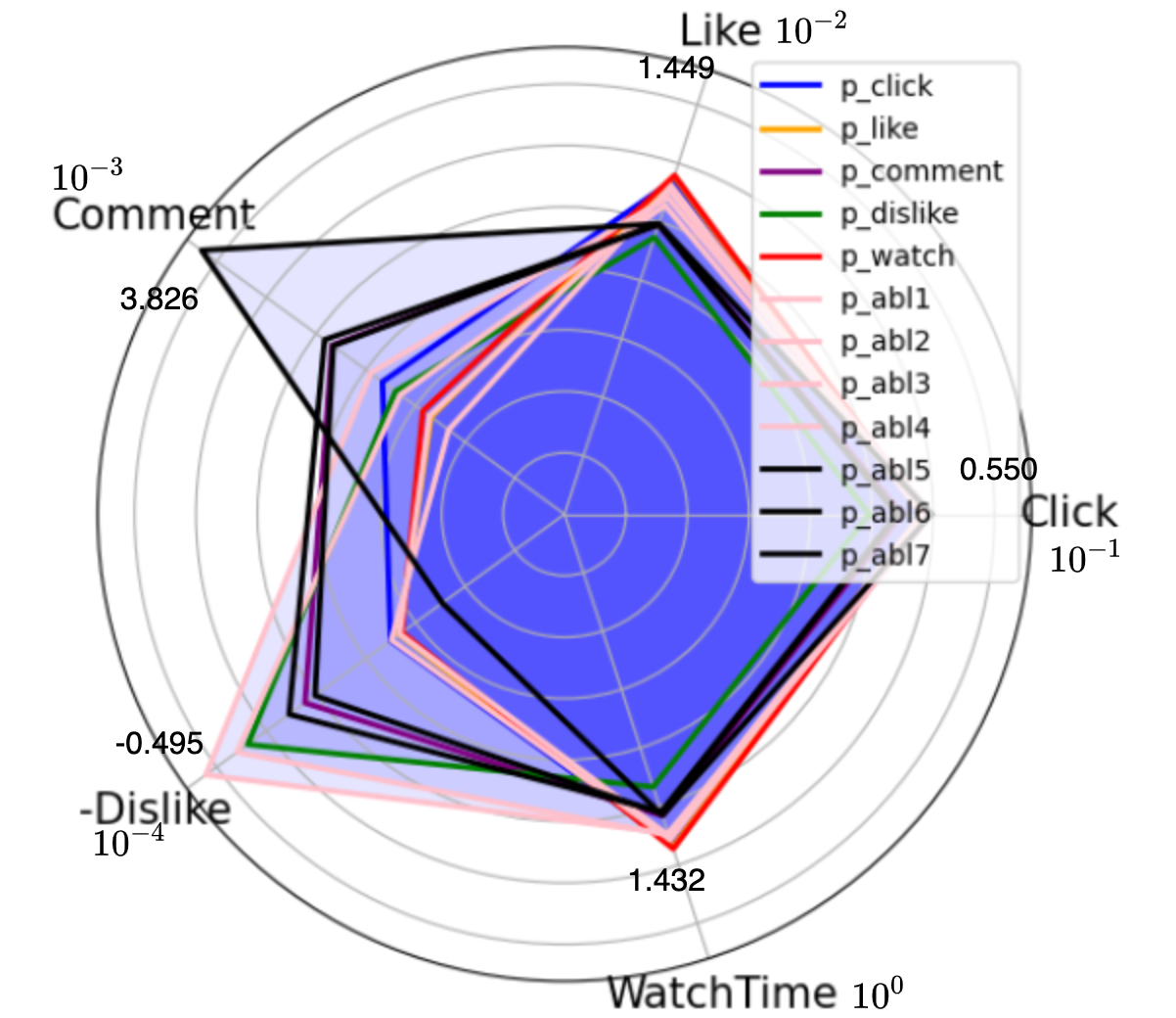}
        \vspace{1pt}
        \caption{\policy Pareto Exploration.}
    \label{fig: add_exp_figa}
    \end{subfigure}%
    ~
    \begin{subfigure}[t]{0.25\textwidth}
        \centering
        \includegraphics[trim=0 50 0 50, width=\textwidth]{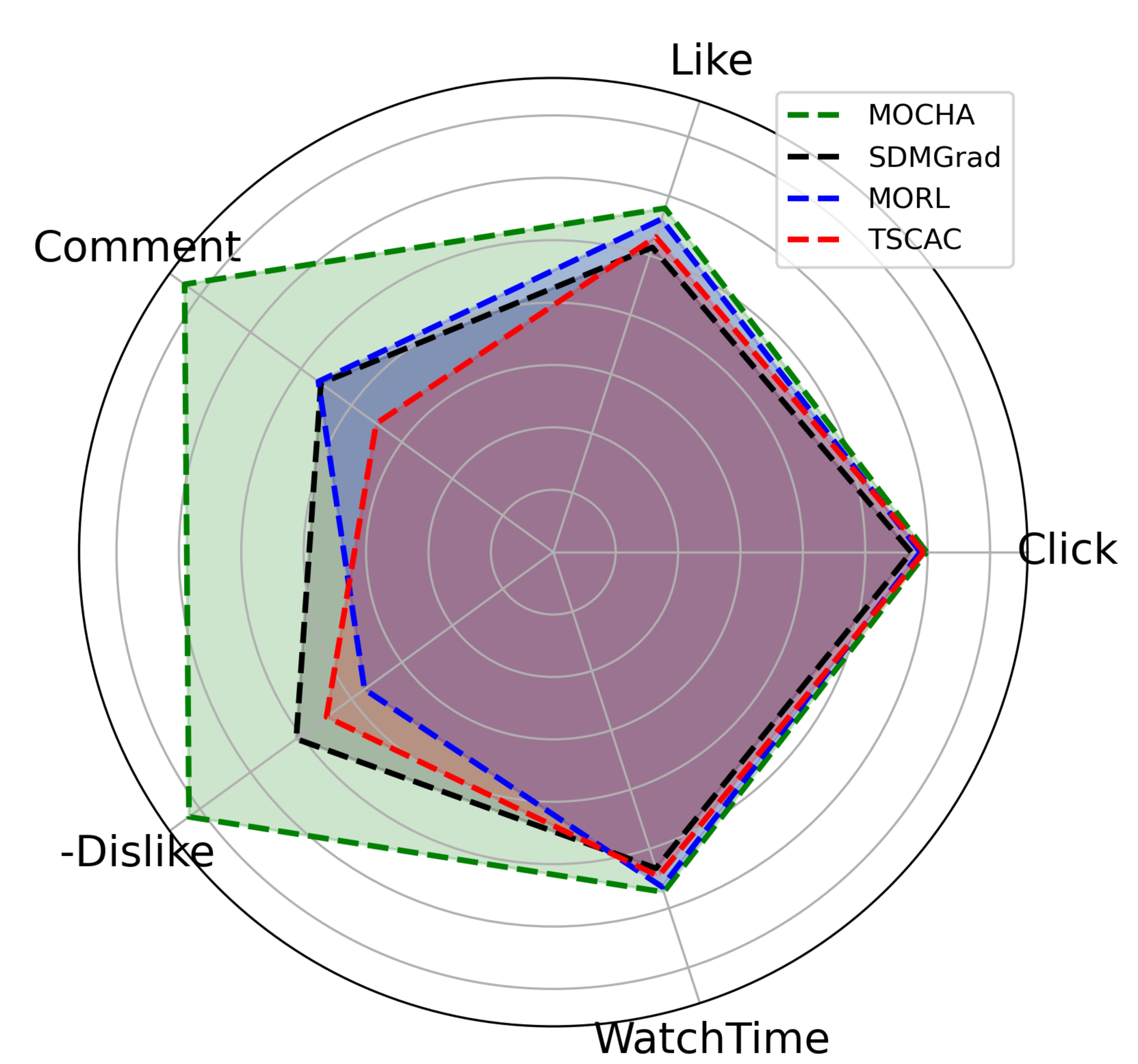}
        \vspace{1pt}
        \caption{Pareto Footprints}
    \label{fig: add_exp_figb}
    \end{subfigure}%
    \caption{\policy and SDMGrad with ablation weight vectors.}
\label{fig: add_exp_fig}
\end{figure}

\textbf{4) Pareto Stationarity Exploration:} 
Here, we provide empirical results for \policy under varying weight vectors $\p$. Specifically, in addition to the 5 one-hot vectors, we have chosen additional weight vectors as in Table \ref{Tab: var_p}. The corresponding results in radar chart are provided in Figure \ref{fig: add_exp_fig}. In Figure \ref{fig: add_exp_figa}, we show the Pareto solutions for \policy explored by the 7 ablation $\p$ vectors in addition to those from the one-hot vectors. In Figure \ref{fig: add_exp_figb}, we further compare the exploration footprints among baseline approaches that include the ablation $\p$ vectors.

From Figure \ref{fig: add_exp_figa}, we can see that with ablation weight vectors $\p$, \policy is exploring more Pareto stationary solutions compared to \policy with only one-hot vectors. In Figure \ref{fig: add_exp_figb}, it further shows that with more $\p$ vectors, \policy explores even wider Pareto solutions than baseline approaches. This empirically confirms our theoretical prediction as well as strengthens the observation that, with increasing number of weight vectors $\p$, \policy possess the potential to explore more Pareto solutions.

\begin{table}
\begin{center}
\caption{Ablation Weight Vectors $\p$}
\begin{tabular}{||c c c c c c||} 
 \hline
 radar result & click & like & comment & dislike & watchtime \\ [0.5ex] 
 \hline\hline
 abl1 & 0.85 & 0.05 & 0.05 & 0 & 0.05 \\ 
 \hline
 abl2 & 0.7 & 0.1 & 0.1 & 0 & 0.1 \\
 \hline
 abl3 & 0.55 & 0.15 & 0.15 & 0 & 0.15 \\
 \hline
 abl4 & 0.4 & 0.2 & 0.2 & 0 & 0.2 \\ 
 \hline
 abl5 & 0.05 & 0.05 & 0.85 & 0.0001 & 0.05 \\
 \hline
 abl6 & 0.10 & 0.10 & 0.70 & 0.0001 & 0.10 \\
 \hline
 abl7 & 0.15 & 0.15 & 0.55 & 0.0001 & 0.15 \\
 \hline
\end{tabular}
\label{Tab: var_p}
\end{center}
\end{table}

%% file: conclusion.tex
\section{Conclusion}
In this paper, we proposed a multi-objective weighted Chebyshev actor-critic (\policyns) algorithm for multi-objective reinforcement learning.
Our proposed \policy method judiciously integrates weighted Chebyshev and actor-critic framework to facilitate systematic Pareto-stationary solution exploration with provable finite-time sample complexity guarantee.
Our numerical experiments with real-world datasets also verified the theoretical results of our \policy method and its practical effectiveness.